\documentclass{article}

\usepackage[final, nonatbib]{nips_2018}
\usepackage[utf8]{inputenc} 
\usepackage[T1]{fontenc}   
\usepackage{hyperref}       
\usepackage{url}            
\usepackage{booktabs}       
\usepackage{amsfonts}       
\usepackage{nicefrac}       
\usepackage{microtype}  
\usepackage{graphicx}
\usepackage{algorithm}
\usepackage{algorithmic}
\usepackage{subfigure}
\usepackage{float}
\usepackage{cite}
\usepackage{amsmath}
\usepackage{comment}
\usepackage{caption}
\usepackage{multicol}

\usepackage{enumitem}
\usepackage[toc,page]{appendix}
\setlist[itemize]{leftmargin=*}

\title{NeST: A Neural Network Synthesis Tool Based on a Grow-and-Prune Paradigm}

\author{
  Xiaoliang Dai\\
  Princeton University\\
  \texttt{xdai@princeton.edu} \\
   \And
   Hongxu Yin \\
   Princeton University\\
  \texttt{hongxuy@princeton.edu} \\
   \And
   Niraj K. Jha \\
   Princeton University\\
  \texttt{jha@princeton.edu} \\
}
\begin{document}

\maketitle

\begin{abstract}
Deep neural networks (DNNs) have begun to have a pervasive impact on various 
applications of machine learning.  However, the problem of finding an optimal 
DNN architecture for large applications is challenging. 
Common approaches go for deeper and larger DNN architectures but may incur 
substantial redundancy. To address these problems, we introduce a network 
growth algorithm that complements network pruning to learn both weights and 
compact DNN architectures during training. We propose a DNN synthesis tool 
(NeST) that combines both methods to automate the generation of compact and 
accurate DNNs. NeST starts with a randomly initialized sparse network called 
the seed architecture. It iteratively tunes the architecture with 
gradient-based growth and magnitude-based pruning of neurons and connections. 
Our experimental results show that NeST yields accurate, yet very
compact DNNs, with a wide range of seed architecture selection. For the
LeNet-300-100 (LeNet-5) architecture, we reduce network parameters by 
$70.2\times$ ($74.3\times$) and floating-point operations (FLOPs) by 
$79.4\times$ ($43.7\times$). For the AlexNet and VGG-16 architectures, we 
reduce network parameters (FLOPs) by $15.7\times$ ($4.6\times$) and 
$30.2\times$ ($8.6\times$), respectively. NeST's grow-and-prune paradigm 
delivers significant additional parameter and FLOPs reduction 
relative to pruning-only methods.

\end{abstract}

\section{Introduction}
\vspace{-1mm}
\footnotetext{This work was supported by NSF Grant No. CNS-1617640.}
Over the last decade, deep neural networks (DNNs) have begun to revolutionize 
myriad research domains, such as computer vision, speech recognition, and 
machine translation~\cite{speechlstm, translation,
human_performance}. Their ability to distill intelligence from a dataset 
through multi-level abstraction can even lead to super-human 
performance~\cite{lecun2015deep}. Thus, DNNs are emerging as a new cornerstone 
of modern artificial intelligence.

Though critically important, how to efficiently derive an appropriate DNN 
architecture from large datasets has remained an open problem. Researchers 
have traditionally derived the DNN architecture by sweeping through its 
architectural parameters and training the corresponding architecture until the 
point of diminishing returns in its performance. This suffers from three 
major problems. First, the widely used back-propagation (BP) algorithm assumes 
a fixed DNN architecture and only trains weights.  Thus, training cannot 
improve the architecture. Second, a trial-and-error methodology can be 
inefficient when DNNs get deeper and contain millions of parameters. Third, 
simply going deeper and larger may lead to large, accurate, but 
over-parameterized DNNs. For example, Han et al. \cite{PruningHS} showed that 
the number of parameters in VGG-16 can be reduced by $13\times$ with no loss 
of accuracy. 

To address these problems, we propose a DNN synthesis tool (NeST) that trains 
both DNN weights and architectures. NeST is inspired by the learning mechanism 
of the human brain, where the number of synaptic connections increases 
upon the birth of a baby, peaks after a few months, and decreases steadily 
thereafter~\cite{spectrum}. NeST starts DNN synthesis from a seed DNN 
architecture (\textit{birth point}). It allows the DNN to \textbf{{grow}} 
connections and neurons based on gradient information (\textit{baby brain}) 
so that the DNN can adapt to the problem at hand. Then, it \textbf{{prunes}} 
away insignificant connections and neurons based on magnitude information 
(\textit{adult brain}) to avoid redundancy. A combination of network growth 
and pruning algorithms enables NeST to generate accurate and compact DNNs.  
We used NeST to synthesize various compact DNNs for the MNIST~\cite{LeNet}
%\footnote{The MNIST dataset contains 70K handwritten digits (60K for training and 10K for testing).}
and ImageNet~\cite{imageNetdataset}
%\footnote{ILSVRC-2012 image classification dataset contains 1.2 million instances for training and 50K for validation.}
datasets.  NeST leads to drastic reductions in the number of parameters
and floating-point operations (FLOPs) relative to the DNN baselines, with 
no accuracy loss.

\vspace{-1mm}
\section{Related Work}
\vspace{-1mm}
%\subsection {Evolutionary Algorithm}
An evolutionary algorithm provides a promising solution to DNN architecture 
selection through evolution of network architectures.
Its search mechanism involves iterations over mutation, 
recombination, and most importantly, evaluation and selection of network 
architectures~\cite{GoogleEvolve, neat}. Additional performance enhancement 
techniques include better encoding methods~\cite{rl4} and algorithmic 
redesign for DNNs~\cite{neat2}.  All these assist with more efficient 
search in the wide DNN architecture space.

Reinforcement learning (RL) has emerged as a new powerful tool to solve this
problem~\cite{rl, rl3, rl2, baidurl}. Zoph et al.~\cite{rl} use a recurrent neural network controller to 
iteratively generate groups of candidate networks, whose performance is then 
used as a reward for enhancing the controller. 
Baker et al.~\cite{rl3} propose a Q-learning based RL approach that 
enables convolutional architecture search. A recent work~\cite{rl2} proposes 
the NASNet architecture that uses RL to search for architectural building 
blocks and achieves better performance than human-invented architectures.

The structure adaptation (SA) approach exploits network \textit{clues} 
(e.g., distribution of weights) to incorporate architecture selection into 
the training process. Existing SA methods can be further divided into two 
categories: constructive and destructive. A constructive 
approach starts with a small network and iteratively 
adds connections/neurons~\cite{DNC, Tiling}. A destructive approach, on the 
other hand, starts with a large network and iteratively removes 
connections/neurons. This can effectively reduce model redundancy. For 
example, recent pruning methods, such as network pruning~\cite{PruningHS, pruning_work, pruning_work1, net_trim}, 
layer-wise surgeon~\cite{layerwise}, sparsity 
learning~\cite{nips3, sparseconv, pruning_work2, pruning_work3}, and dynamic network 
surgery~\cite{nips2}, can offer extreme compactness for existing 
DNNs with no or little accuracy loss.

\vspace{-1mm}
\section{Synthesis Methodology}
\vspace{-1mm}
In this section, we propose NeST that leverages both constructive and 
destructive SA approaches through a grow-and-prune paradigm. Unless otherwise 
stated, we adopt the notations given in Table~\ref{notation-table} to 
represent various variables.

\vspace{-5mm}
\begin{table}[h]
\footnotesize
\centering
\caption{Notations and descriptions}
\label{notation-table}
\begin{tabular}{lr|lr}
\hline
Label & Description & Label & Description\\
\hline
$L$ & DNN loss function & $\textbf{W}^{l}$ & weights between $(l-1)^{th}$ and $l^{th}$ layer\\
 $x^{l}_{n}$ & output value of $n^{th}$ neuron in $l^{th}$ layer & $\textbf{b}^{l}$ & biases in $l^{th}$ layer\\
 $u^{l}_{m}$ & input value of $m^{th}$ neuron in $l^{th}$ layer& $\textbf{R}^{d_0\times d_1 \times d_2}$ & $d_0$ by $d_1 $ by $d_2$ matrix with real elements\\
\hline
\end{tabular}
\vskip -0.1in
\end{table}

\begin{figure*}[h]
\begin{center}
\includegraphics[width=125mm]{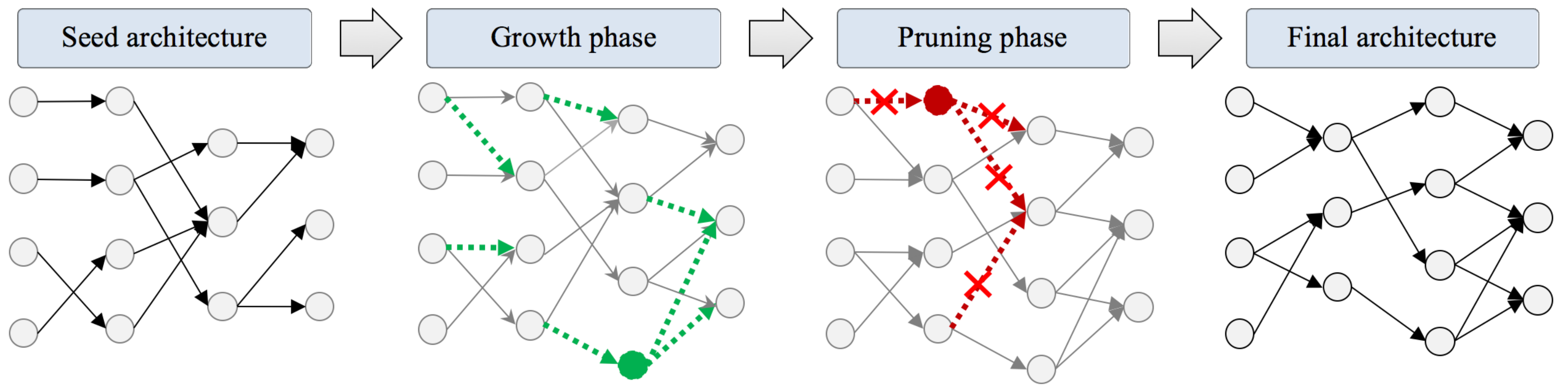}
\end{center}
\caption{An illustration of the architecture synthesis flow in NeST.}
\label{fig:phases}
\end{figure*}

\vspace{-1mm}
\subsection{Neural Network Synthesis Tool}
\vspace{-1mm}
We illustrate the NeST approach with Fig.~\ref{fig:phases}. Synthesis begins 
with an initial seed architecture, typically initialized as a sparse and 
partially connected DNN. We also ensure that all neurons are connected in 
the seed architecture. Then, NeST utilizes two sequential phases to synthesize 
the DNN: (i) gradient-based growth phase, and (ii) magnitude-based pruning 
phase. In the growth phase, the gradient information in the architecture space 
is used to gradually grow new connections, neurons, and feature maps to 
achieve the desired accuracy. In the pruning phase, the DNN inherits the 
synthesized architecture and weights from the growth phase and iteratively 
removes redundant connections and neurons, based on their magnitudes. Finally, 
NeST comes to rest at a lightweight DNN model that incurs no accuracy 
degradation relative to a fully connected model. 
\vspace{-1mm}
\subsection{Gradient-based Growth}
\vspace{-1mm}
In this section, we explain our algorithms to grow connections, neurons, and 
feature maps. %based on gradient information.
\vspace{-1mm}
\subsubsection{Connection Growth}
\vspace{-1mm}
The connection growth algorithm greedily activates useful, but currently 
`dormant,' connections. We incorporate it in the following learning policy:
       
\noindent
\textbf{Policy 1:} Add a connection $w$ iff it can quickly reduce the value of 
loss function $L$.

\begin{figure}
\begin{minipage}[h]{85mm}
\centering
\includegraphics[width=85mm]{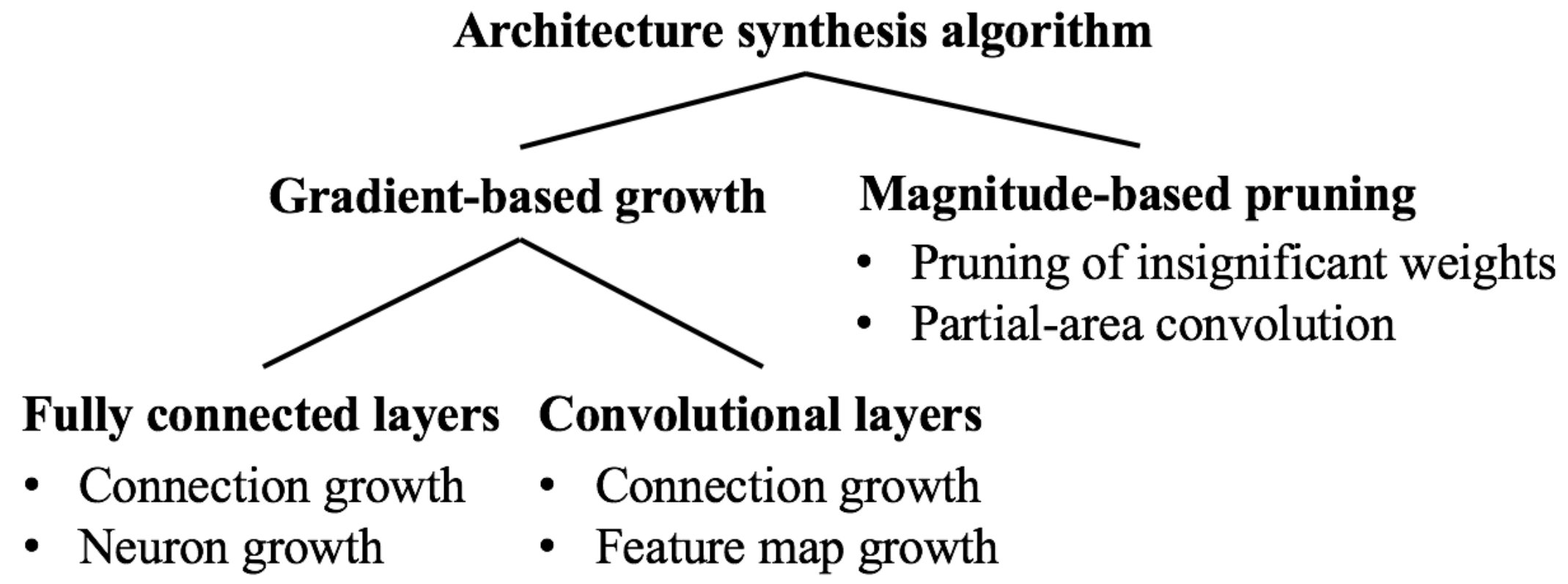}
\caption{Major components of the DNN architecture synthesis algorithm in NeST.}
\label{fig:algo}
\end{minipage}
\begin{minipage}[h]{3mm}
\ 
\end{minipage}
\begin{minipage}[h]{40mm}
\centering
\includegraphics[width=40mm]{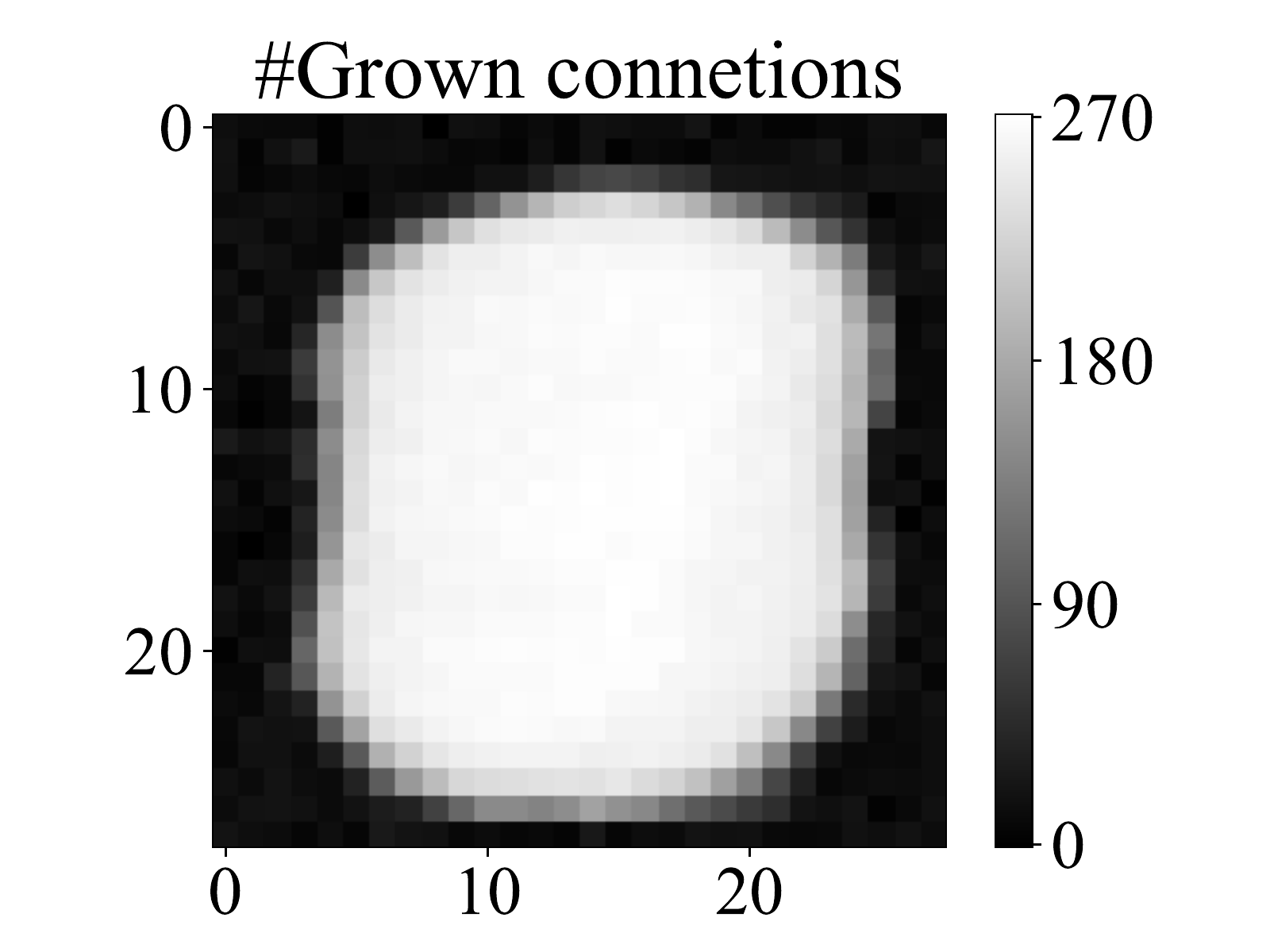}
\caption{Grown connections from the input layer to the first layer of LeNet-300-100.}
\label{fig:mlp_conn_grow}
\end{minipage}
\vspace{-3mm}
\end{figure}

The DNN seed contains only a small fraction of active 
connections to propagate gradients. To locate the `dormant' 
connections that can reduce $L$ effectively, we evaluate 
$\partial L / \partial w$ for all the `dormant' connections $w$
(computed either using the whole training set or a large batch). 
Policy 1 activates `dormant' connections iff they are the most efficient
at reducing $L$. This can also assist with avoiding local minima and achieving 
higher accuracy~\cite{dsd}.  To illustrate this policy, we plot the connections 
grown from the input to the first layer of LeNet-300-100~\cite{LeNet} (for 
the MNIST dataset) in Fig.~\ref{fig:mlp_conn_grow}. The image center has a 
much higher grown density than the margins, consistent with the fact 
that the MNIST digits are centered. 

From a neuroscience perspective, our connection growth algorithm coincides with 
the Hebbian theory: ``Neurons that fire together wire 
together~\cite{HebbRule}."  We define the stimulation magnitude of the $m^{th}$ 
presynaptic neuron in the $(l+1)^{th}$ layer and the $n^{th}$ postsynaptic 
neuron in the $l^{th}$ layer as $\partial L/\partial u^{l+1}_{m}$ 
and $x^{l}_{n}$, respectively.  The connections activated based on 
Hebbian theory would have a strong correlation between presynaptic and 
postsynaptic cells, thus a large value of 
$\left|(\partial L /  \partial u^{l+1}_{m})x^{l}_{n}\right|$. This is also the 
magnitude of the gradient of $L$ with respect to $w$ ($w$ is the weight that 
connects $u^{l+1}_{m}$ and $x^{l}_{n}$):
\begin{equation}
    \left|\partial L /  \partial w\right|=\left|(\partial L /  \partial u^{l+1}_{m})x^{l}_{n}\right|
\end{equation}
Thus, this is mathematically equivalent to Policy 1.

\vspace{-1mm}
\subsubsection{Neuron Growth}
\vspace{-1mm}
Our neuron growth algorithm consists of two steps: (i) connection 
establishment and (ii) weight initialization. The neuron growth policy
is as follows:

\noindent
\textbf{Policy 2:} In the $l^{th}$ layer, add a new neuron as a shared 
intermediate node between existing neuron pairs that have high postsynaptic 
($x$) and presynaptic ($\partial L / \partial u$) neuron correlations (each 
pair contains one neuron from the $(l-1)^{th}$ layer and the other from 
the $(l+1)^{th}$ layer). Initialize weights based on batch gradients to 
reduce the value of $L$.

\begin{algorithm}[t]
\footnotesize
   \caption{Neuron growth in the $l^{th}$ layer}
   \label{alg:neuron_grow}
\begin{algorithmic}
   \STATE {\bfseries Input:} $\alpha$ - birth strength, $\beta$ - growth ratio
   \STATE {\bfseries Denote:} $M$ - number of neurons in the $(l+1)^{th}$ 
   layer, $N$ - number of neurons in the $(l-1)^{th}$ layer, $\textbf{G} 
   \in R^{M\times N}$ - bridging gradient matrix, $avg$ - extracts mean value of non-zero elements
   \STATE Add a neuron in the $l^{th}$ layer, initialize $\textbf{w}^{out} = \vec{\textbf{0}}\in R^{M}$, $\textbf{w}^{in} = \vec{\textbf{0}}\in R^{N}$
   \FOR {$1\leq m\leq M, 1\leq n\leq N$}
   \STATE $G_{m,n} = \frac{\partial L}{ \partial u^{l+1}_{m}} \times x^{l-1}_{n}$ 
   \ENDFOR
   \STATE $thres$ = $(\beta MN)^{th}$ largest element in $abs(\textbf{G})$
   \FOR {$1\leq m\leq M, 1\leq n\leq N$}
    \IF {$|G_{m,n}| > thres$}
     \STATE $\delta w = \sqrt{|G_{m,n}|}$ $\times rand\{1, -1\}$
     \STATE $w^{out}_{m} \leftarrow w^{out}_{m} +\delta w$,\ $w^{in}_{n} \leftarrow w^{in}_{n} + \delta w \times sgn(G_{m,n})$
    \ENDIF
    \STATE $\textbf{w}^{out} \leftarrow \textbf{w}^{out} \times \alpha \frac{avg(abs(\textbf{W}^{l+1}))}{avg(abs(\textbf{w}^{out}))}$
    ,\ $\textbf{w}^{in} \leftarrow \textbf{w}^{in} \times 
    \alpha \frac{avg(abs(\textbf{W}^{l}))}{avg(abs(\textbf{w}^{in}))}$
   \ENDFOR
   \STATE Concatenate network weights $\textbf{W}$ with $\textbf{w}^{in}$, $\textbf{w}^{out}$
\end{algorithmic}
\end{algorithm}

Algorithm \ref{alg:neuron_grow} incorporates Policy 2 and illustrates the 
neuron growth iterations in detail. Before adding a neuron to the 
$l^{th}$ layer, we evaluate the bridging gradient between the neurons at 
the previous $(l-1)^{th}$ and subsequent $(l+1)^{th}$ layers. We connect the 
top $\beta \times 100\%$ ($\beta$ is the growth ratio) correlated neuron pairs
through a new neuron in the $l^{th}$ layer. We initialize the weights based on 
the bridging gradient to enable gradient descent, thus decreasing
the value of $L$. 

We implement a square root rule for weight initialization to 
imitate a BP update on the
bridging connection $w_{b}$, which connects $x_{n}^{l-1}$ 
%[output value of the $n^{th}$ neuron in the $(l-1)^{th}$ layer]
and $u_{m}^{l+1}$.
%[input value of the $m^{th}$ neuron in the $(l+1)^{th}$ layer]
The BP update 
leads to a change in $u_{m}^{l+1}$:

\vspace*{-1mm}
\begin{equation}
|\Delta u_{m}^{l+1}|_{b.p.} = |x_{n}^{l-1} \times \delta w_{b}| = \eta|x_{n}^{l-1} \times G_{m, n}|
\end{equation}

\vspace*{-1mm}
where $\eta$ is the learning rate.  In Algorithm 1, when we connect the newly 
added neuron (in the $l^{th}$ layer) with $x_{n}^{l-1}$ and $u_{m}^{l+1}$, we 
initialize their weights to the square root of the magnitude of the bridging 
gradient:

\vspace*{-1mm}
\begin {equation}
|\delta w^{in}_{n}| = |\delta w^{out}_{m}|  =  \sqrt{|G_{m,n}|}
\end{equation}

\vspace*{-1mm}
where $ \delta w^{in}_{n}$ ($\delta w^{out}_{m}$) is the initialized value of the weight that connects the newly added neuron 
with $x_{n}^{l-1}$ ($u_{m}^{l+1}$).  The weight initialization rule 
leads to a change in $u_{m}^{l+1}$:

\vspace*{-1mm}
\begin{equation}
|\Delta u_{m}^{l+1}| = |f (x_{n}^{l-1} \times \delta w^{in}_{n}) \times \delta w^{out}_{m}|
\end{equation}

\vspace*{-1mm}
where $f$ is the neuron activation function. Suppose $tanh$ 
is the activation function. Then,

\vspace*{-1mm}
\begin{equation}
f(x)=tanh(x)\approx x, \text{if} \ x \ll 1
\end{equation}

\vspace*{-1mm}
Since $\delta w^{in}_{n}$ and $\delta w^{out}_{m}$ are typically very small, 
the approximation in Eq. (5) leads to Eq. (6).

\vspace*{-1mm}
\begin{equation}
|\Delta u_{m}^{l+1}| \approx |x_{n}^{l-1}\times \delta w^{in}_{n} \times \delta w^{out}_{m}| = \frac{1}{\eta} \times |\Delta u_{m}^{l+1}|_{b.p.}\\
\end{equation}

\vspace*{-1mm}
This is linearly proportional to the effect of a BP update. 
Thus, our weight initialization mathematically imitates a BP 
update.  Though we illustrated the algorithm with the $tanh$ activation
function, the weight initialization rule works equally well with other 
activation functions, such as rectified linear unit (ReLU)
%$^{1}$\footnote{1. ReLU: $f(x) = max(0, x)$})
and leaky rectified linear unit (Leaky ReLU).
%$^{2}$\footnote{2. Leaky ReLU: $f(x) = max(0.01x, x)$}).

We use a birth strength factor $\alpha$ to strengthen the connections 
of a newly grown neuron.  This prevents these connections from 
becoming too weak to survive the pruning phase.  
Specifically, after square root rule based weight initialization, we scale 
up the newly added weights by
\begin{equation}
\textbf{w}^{out} \leftarrow \alpha \textbf{w}^{out} \times \frac{avg(abs(\textbf{W}^{l+1}))}{avg(abs(\textbf{w}^{out}))}, \ \ 
\textbf{w}^{in} \leftarrow \alpha \textbf{w}^{in} \times \frac{avg(abs(\textbf{W}^{l}))}{avg(abs(\textbf{w}^{in}))}
\end{equation}
where $avg$ is an operation that extracts the mean value of all non-zero 
elements.  This strengthens new weights.  In practice, we find $\alpha>0.3$ 
to be an appropriate range.

\subsubsection{Growth in the Convolutional Layers}
\vspace{-1mm}
Convolutional layers share the connection growth methodology of Policy 1.  
However, instead of neuron growth, we use a unique feature map growth algorithm for convolutional
layers.  In a convolutional layer, we convolve input images with kernels to generate 
feature maps.  Thus, to add a feature map, we need to initialize the 
corresponding set of kernels. We summarize the feature map growth policy as 
follows:

\noindent
\textbf{Policy 3:} To add a new feature map to the convolutional layers, 
randomly generate sets of kernels, and pick the set of kernels that reduces 
$L$ the most.

In our experiment, we observe that the percentage reduction in $L$ for Policy 
3 is approximately twice as in the case of the naive approach that initializes the 
new kernels with random values. 

\vspace{-1mm}
\subsection{Magnitude-based Pruning}
\vspace{-1mm}
We prune away insignificant connections and neurons based on the magnitude 
of weights and outputs:

\noindent
\textbf{Policy 4:} Remove a connection (neuron) iff the magnitude of the 
weight (neuron output) is smaller than a pre-defined threshold.

We next explain two variants of Policy 4: pruning of insignificant weights and 
partial-area convolution.

\vspace{-1mm}
\subsubsection{Pruning of Insignificant Weights}
\vspace{-1mm}
Han et al.~\cite{PruningHS} show that magnitude-based pruning can successfully cut down the memory and computational costs. We extend this approach to incorporate the batch 
normalization technique.  Such a technique can reduce the internal covariate 
shift by normalizing layer inputs and improve the training 
speed and behavior. Thus, it has been widely applied to large DNNs~\cite{bn}.  
Consider the $l^{th}$ batch normalization layer:
\vspace{-1mm}
\begin{equation}
\textbf{u}^{l} = [(\textbf{W}^{l}\textbf{x}^{l-1} + \textbf{b}^{l}) - \textbf{E}] \oslash{\textbf{V}}
= \textbf{W}_{*}^{l} \textbf{x} + \textbf{b}_{*}^{l}
\end{equation}
where $\textbf{E}$ and $\textbf{V}$ are batch normalization terms, 
and $\oslash$ depicts the Hadamard (element-wise) division operator. We 
define effective weights $\textbf{W}_{*}^{l}$ and effective biases $\textbf{b}_{*}^{l}$ as:
\vspace{-0.5mm}
\begin{equation}
\textbf{W}_{*}^{l} = \textbf{W}^{l}\oslash{\textbf{V}}, \textbf{b}_{*}^{l} = (\textbf{b}^{l} - \textbf{E})\oslash{\textbf{V}}
\end{equation}

\vspace*{-1mm}
We treat connections with small effective weights as insignificant.  Pruning 
of insignificant weights is an iterative process. In each iteration, we only 
prune the most insignificant weights (e.g., top 1\%) for each layer, 
and then retrain the whole DNN to recover its performance.
 
\vspace{-1mm}
\subsubsection{Partial-area Convolution}
\vspace{-1mm}
In common convolutional neural networks (CNNs), the convolutional layers 
typically consume $\sim5\%$ of the parameters, but contribute 
to $\sim90$-$95\%$ of the total FLOPs~\cite{oneweirdtrick}. 
In a convolutional layer, kernels shift and 
convolve with the entire input image.  This process 
incurs redundancy, since not the whole input image is of interest 
to a particular kernel.  Anwar et al.~\cite{featuremappruning} presented a method to prune all 
connections from a not-of-interest input image to a particular 
kernel. This method reduces FLOPs but incurs performance degradation~\cite{featuremappruning}.

\begin{figure}
\begin{minipage}[t]{58mm}
\centering
\footnotesize
\includegraphics[width=40mm]{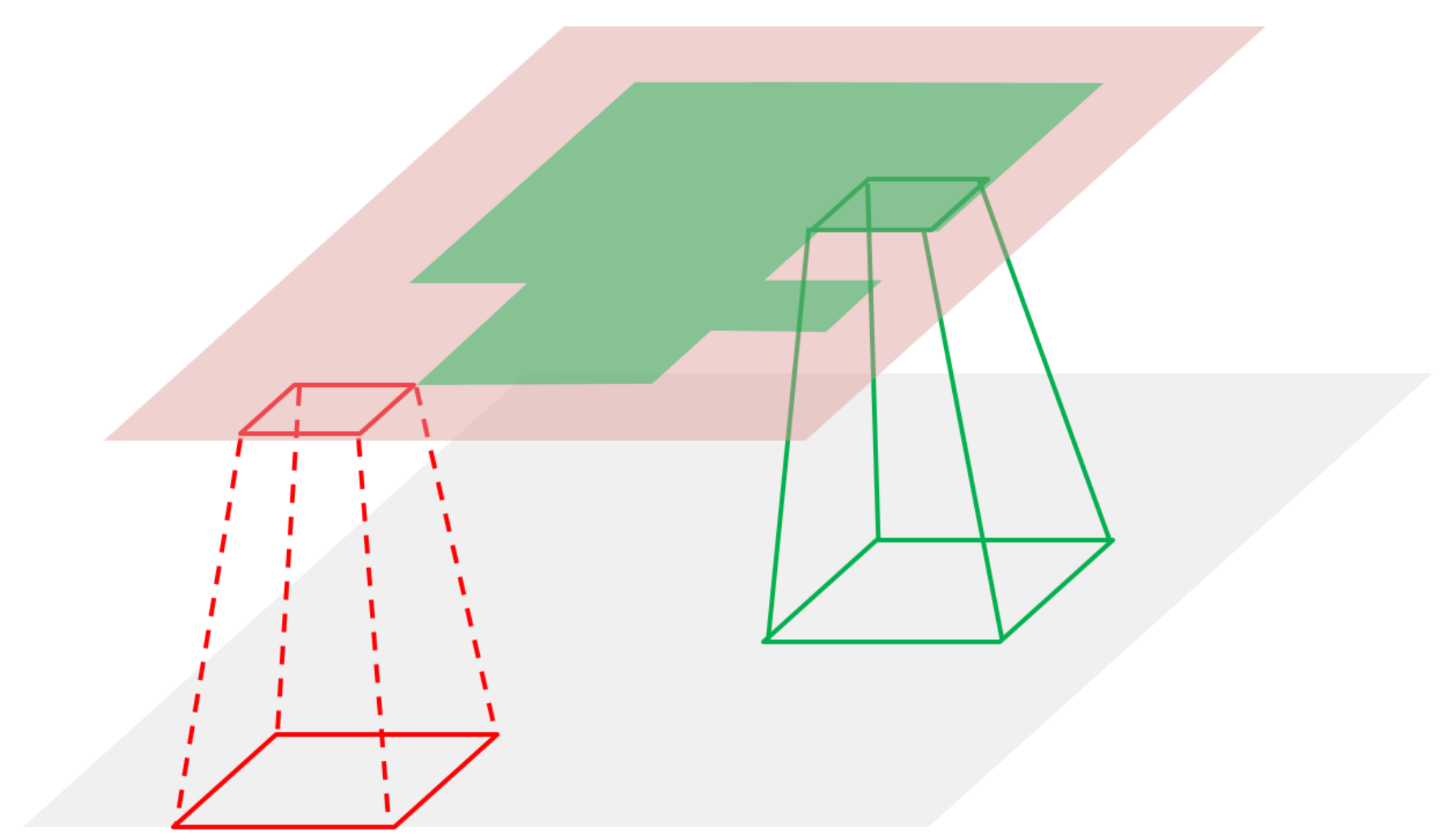}
\caption{Pruned connections (dashed red lines) and remaining connections (solid green lines) in partial-area convolution.}
\label{fig:part_conv}
\end{minipage}
\begin{minipage}[t]{3mm}
\ 
\end{minipage}
\begin{minipage}[t]{75mm}
\centering
\includegraphics[width=75mm]{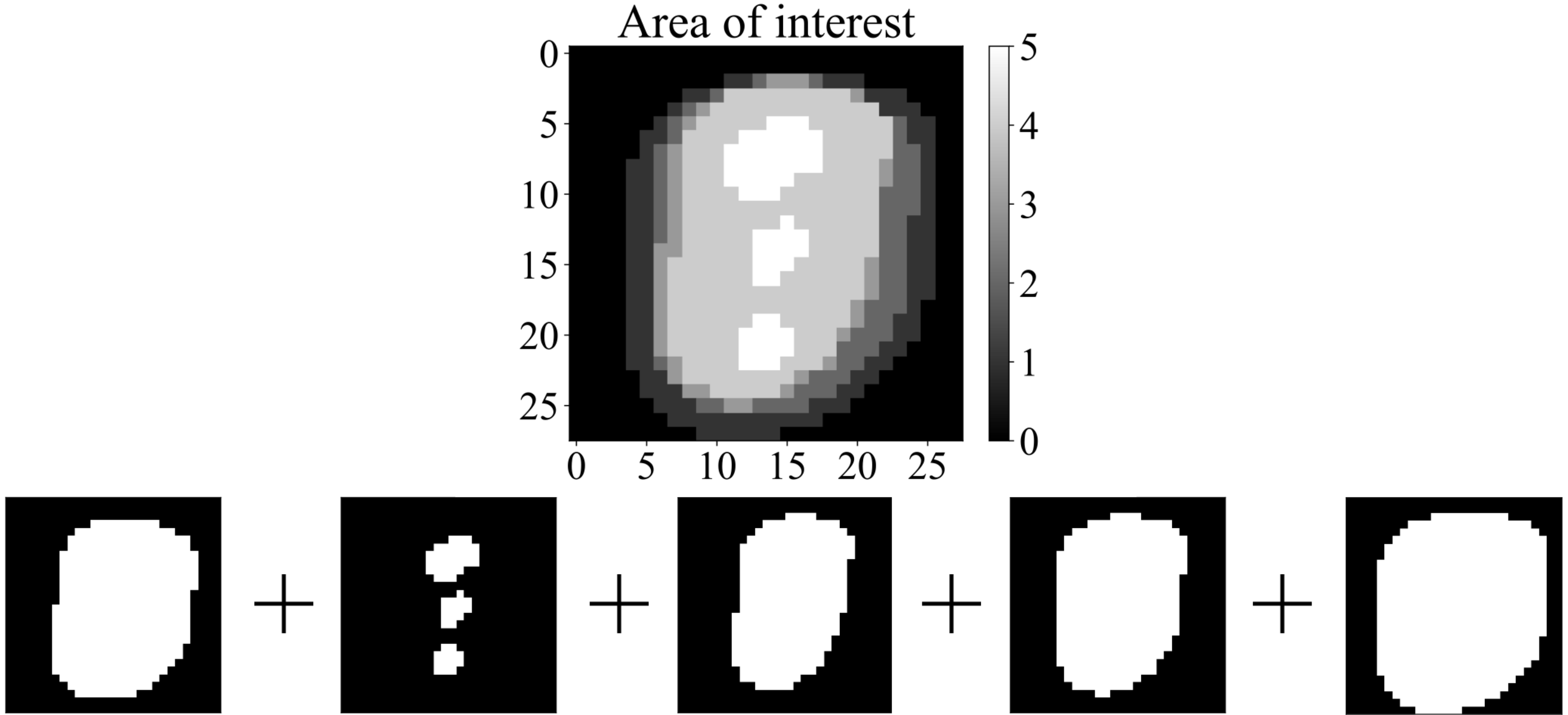}
\caption{Area-of-interest for five different kernels in the first layer of 
LeNet-5.}
\label{fig:conv_freq}
\end{minipage}
\vspace{-3mm}
\end{figure}

Instead of discarding an entire image, our proposed partial-area convolution 
algorithm allows kernels to convolve 
with the image areas that are of interest. We refer to such an area 
as \textit{\textbf{area-of-interest}}. We prune connections to other 
image areas. We illustrate this process 
in Fig.~\ref{fig:part_conv}. The green area depicts 
\textit{\textbf{area-of-interest}}, whereas the red area 
depicts parts that are not of interest. Thus, green connections (solid 
lines) are kept, whereas red ones (dashed lines) are pruned away.

Partial-area convolution pruning is an iterative process.  We present one iteration in Algorithm~\ref{alg:pac}.  We first convolve 
$M$ input images with $M\times N$ convolution kernels and generate 
$M\times N$ feature maps, which are stored in a four-dimensional 
feature map matrix $\textbf{C}$.  We set the pruning threshold $thres$ to
the $(100\gamma) ^{th}$ percentile of all elements in $abs(\textbf{C})$, where $\gamma$ is the pruning ratio, typically 1\% in our experiment. We mark the elements whose 
values are below this threshold as insignificant, and prune away their 
input connections.  We retrain the 
whole DNN after each pruning iteration. In our current implementation, we utilize a mask $\textbf{Msk}$ to disregard the pruned convolution area. 

\begin{algorithm}[tb]
   \caption{Partial-area convolution}
   \label{alg:pac}
\begin{algorithmic}
\footnotesize
   \STATE {\bfseries Input:} $\textbf{I}$ - $M$ input images,
	$\textbf{K}$ - kernel matrix, $\textbf{Msk}$ - feature map mask, 
	$\gamma$ - pruning ratio 
   \STATE {\bfseries Output:} $\textbf{Msk}$, $\textbf{F}$ - $N$ feature maps  
   \STATE {\bfseries Denote:} $\textbf{C} \in R^{M\times N\times P\times Q}$ - Depthwise feature map,
   $\otimes$ - Hadamard  (element-wise)  multiplication
   \FOR {$1\leq m\leq M, 1\leq n\leq N$}
     \STATE $\textbf{C}_{m, n}$ = $convolve$($\textbf{I}_{m}$, $\textbf{K}_{m, n}$)
   \ENDFOR
   \STATE  $thres$ = $(\gamma MNPQ)^{th}$ largest element in $abs(\textbf{C})$
   \FOR {$1\leq m\leq M, 1\leq n\leq N, 1\leq p\leq P, 1\leq q\leq Q$}
   \IF {$|C_{m, n, p, q}| < thres$}
   \STATE $Msk_{m, n, p, q} = 0 $
   \ENDIF
   \ENDFOR
   \STATE $\textbf{C} \leftarrow \textbf{C}  \otimes \textbf{Msk}$,\ \ $\textbf{F} \leftarrow \Sigma_{m=1}^{M} \textbf{C}_{m}$
\end{algorithmic}
\end{algorithm}

Partial-area convolution enables additional FLOPs reduction without any 
performance degradation. For example, we can reduce FLOPs in LeNet-5~\cite{LeNet} by 
$2.09\times$ when applied to MNIST.  
Compared to the conventional CNNs that force a fixed 
square-shaped \textit{\textbf{area-of-interest}} on all kernels, we allow 
each kernel to self-explore the preferred shape of its 
\textit{\textbf{area-of-interest}}. Fig.~\ref{fig:conv_freq} shows the \textit{\textbf{area-of-interest}} 
found by the layer-1 kernels in LeNet-5 when applied to MNIST. We observe 
significant overlaps in the image central area, which most kernels are 
interested in.

\vspace{-1mm}
\section{Experimental Results}
\vspace{-1mm}
We implement NeST using Tensorflow~\cite{tensorflow} and PyTorch~\cite{pytorch} on
Nvidia GTX 1060 and Tesla P100 GPUs. We use NeST to synthesize compact DNNs
for the MNIST and ImageNet datasets. We select DNN seed 
architectures based on clues (e.g., depth, kernel size, etc.) from the existing LeNets, AlexNet, and VGG-16 architectures, respectively.  NeST 
exhibits two major advantages:

\vspace{-3mm}
\begin{itemize}\itemsep-0em 
\item \textbf{Wide seed range}: NeST yields high-performance DNNs with a wide 
range of seed architectures.  Its ability to start from a wide range of 
seed architectures alleviates reliance on human-defined architectures, and 
offers more freedom to DNN designers.
\item \textbf{Drastic redundancy removal}: NeST-generated DNNs are very 
compact. Compared to the DNN architectures
generated with pruning-only methods, DNNs generated through our grow-and-prune 
paradigm have much fewer parameters and require much fewer FLOPs.
\end{itemize}

\vspace{-3mm}
\subsection{LeNets on MNIST}
\vspace{-1mm}
We derive the seed architectures from the original LeNet-300-100 and 
LeNet-5 networks~\cite{LeNet}.  LeNet-300-100 is a multi-layer perceptron
with two hidden layers.  LeNet-5 is a CNN with two convolutional layers 
and three fully connected layers.  We use the affine-distorted MNIST 
dataset~\cite{LeNet}, on which LeNet-300-100 (LeNet-5) can achieve an error 
rate of $1.3\%$ ($0.8\%$).  We discuss our results next.

\vspace{-1mm}
\subsubsection{Growth Phase}
\vspace{-1mm}
First, we derive nine (four) seed architectures for LeNet-300-100 
(LeNet-5).  These seeds contain fewer neurons and connections 
per layer than the original LeNets.  The number of neurons in each layer is 
the product of a ratio $r$ and the corresponding number in the original 
LeNets (e.g., the seed architecture for LeNet-300-100 becomes LeNet-120-40 
if $r=0.4$).  We randomly initialize only 10\% of all possible connections 
in the seed architecture. Also, we ensure that all neurons in the network are 
connected.

We first sweep $r$ for LeNet-300-100 (LeNet-5) from 0.2 (0.5) to 1.0 (1.0) 
with a step-size of 0.1 (0.17), and then grow the DNN architectures from
these seeds. We study the impact of these seeds on the GPU time for growth 
and post-growth DNN sizes under the same target accuracy (this accuracy is 
typically a reference value for the architecture). We summarize the results 
for LeNets in Fig.~\ref{fig:LeNet_grow}. We have two interesting findings for 
the growth phase:
%We can make two major observations for the growth phase:

\begin{figure}
\begin{center}
\begin {tabular}{ccc}
\includegraphics[width=65mm]{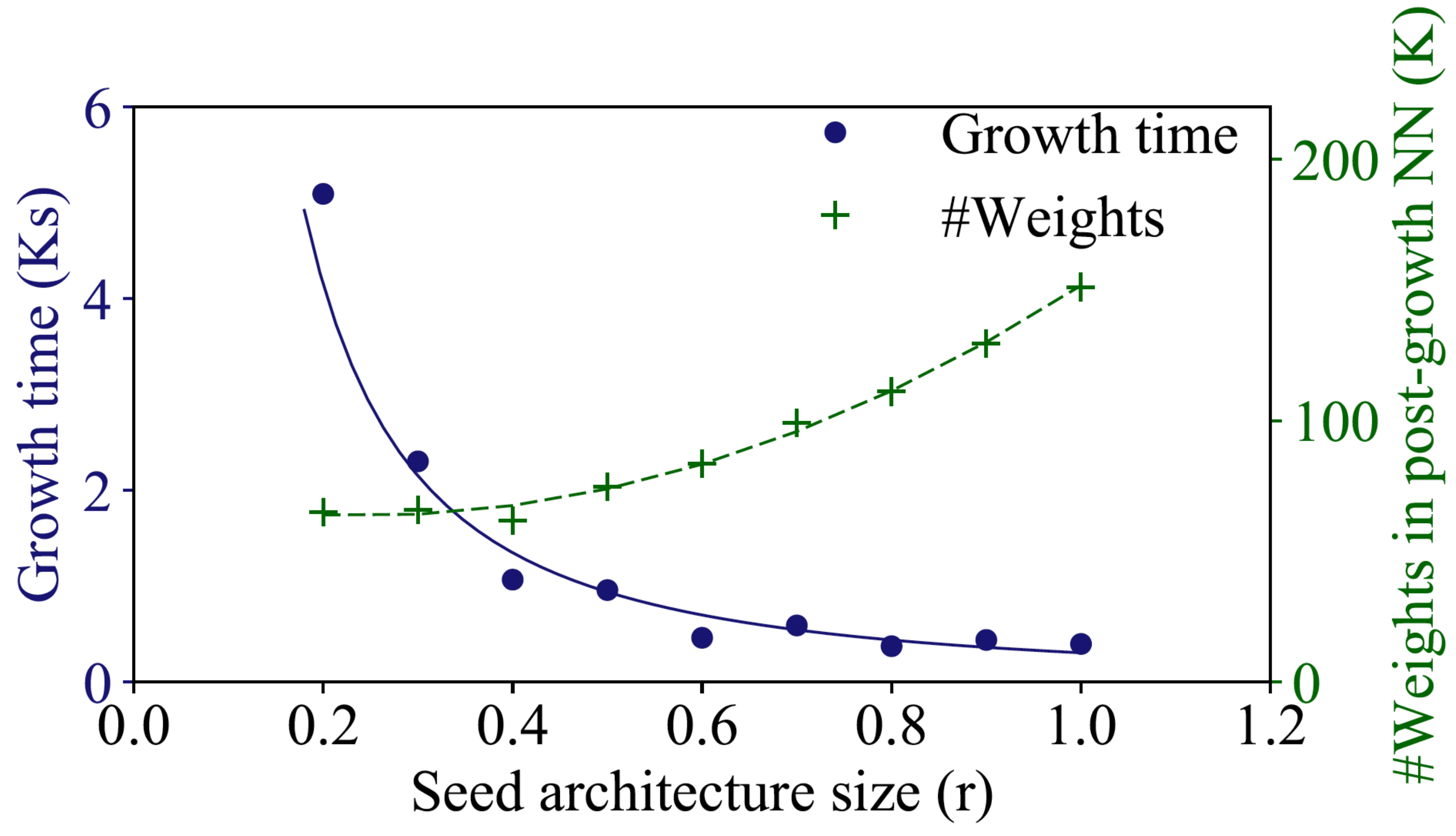}
&  
\includegraphics[width=65mm]{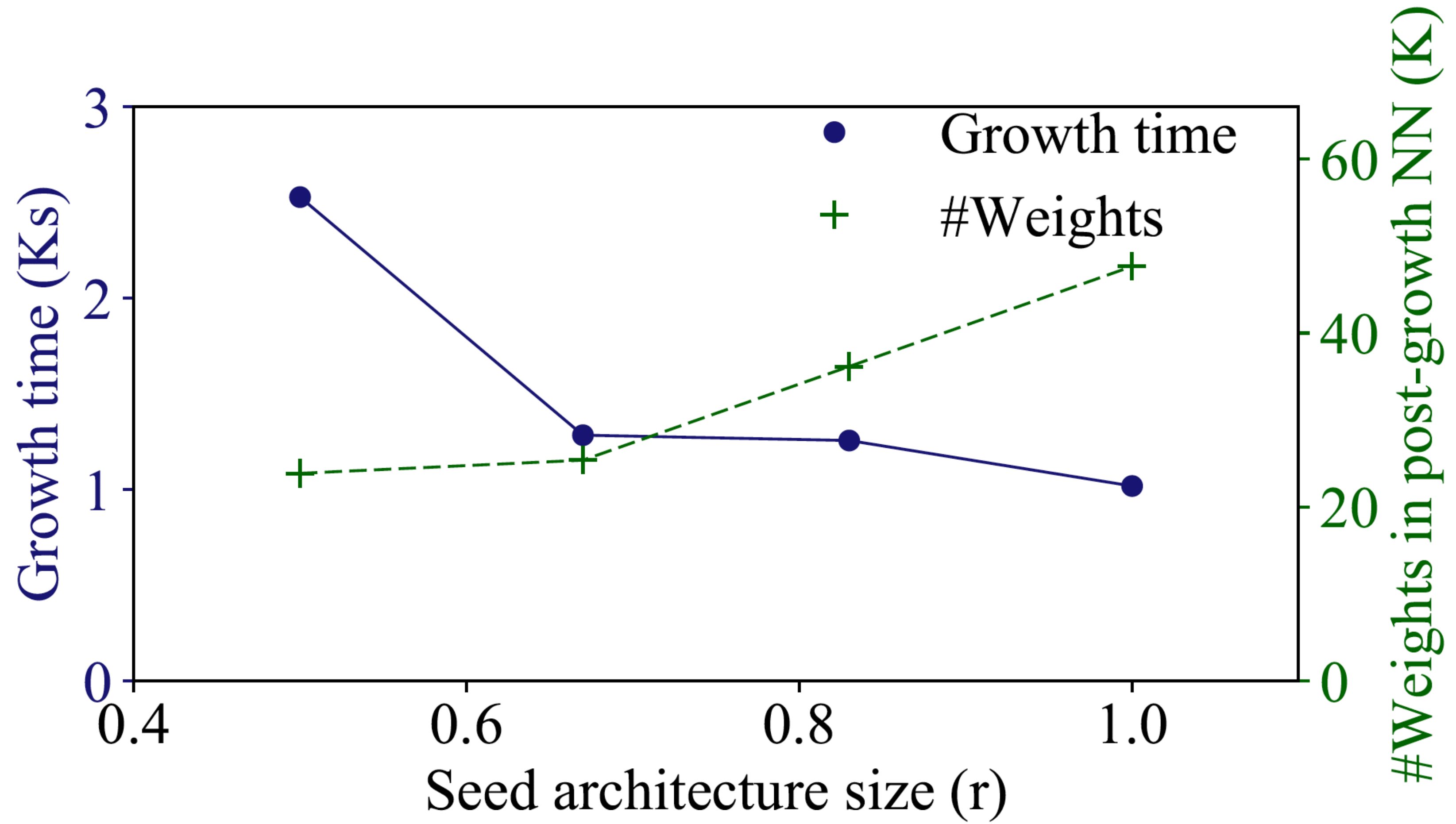}
\end {tabular}
\end {center}
\vspace{-1mm}
\caption{Growth time vs. post-growth DNN size trade-off for various seed 
architectures for LeNet-300-100 (left) and LeNet-5 (right) to achieve a
1.3\% and 0.8\% error rate, respectively.}
\label{fig:LeNet_grow}
\end{figure}

\begin{figure}
\begin{center}
\begin {tabular}{ccc}
\includegraphics[width=65mm]{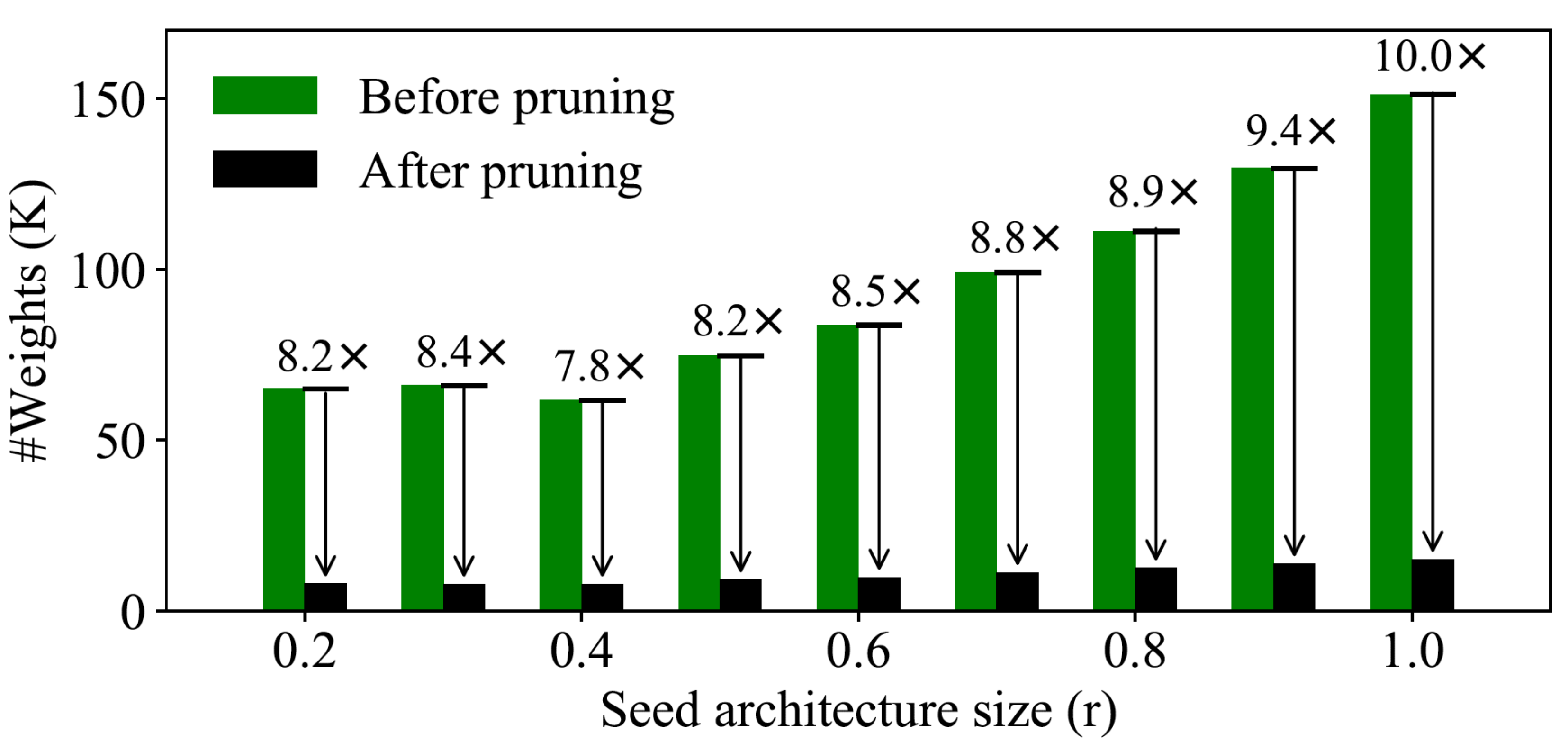}
&  
\includegraphics[width=65mm]{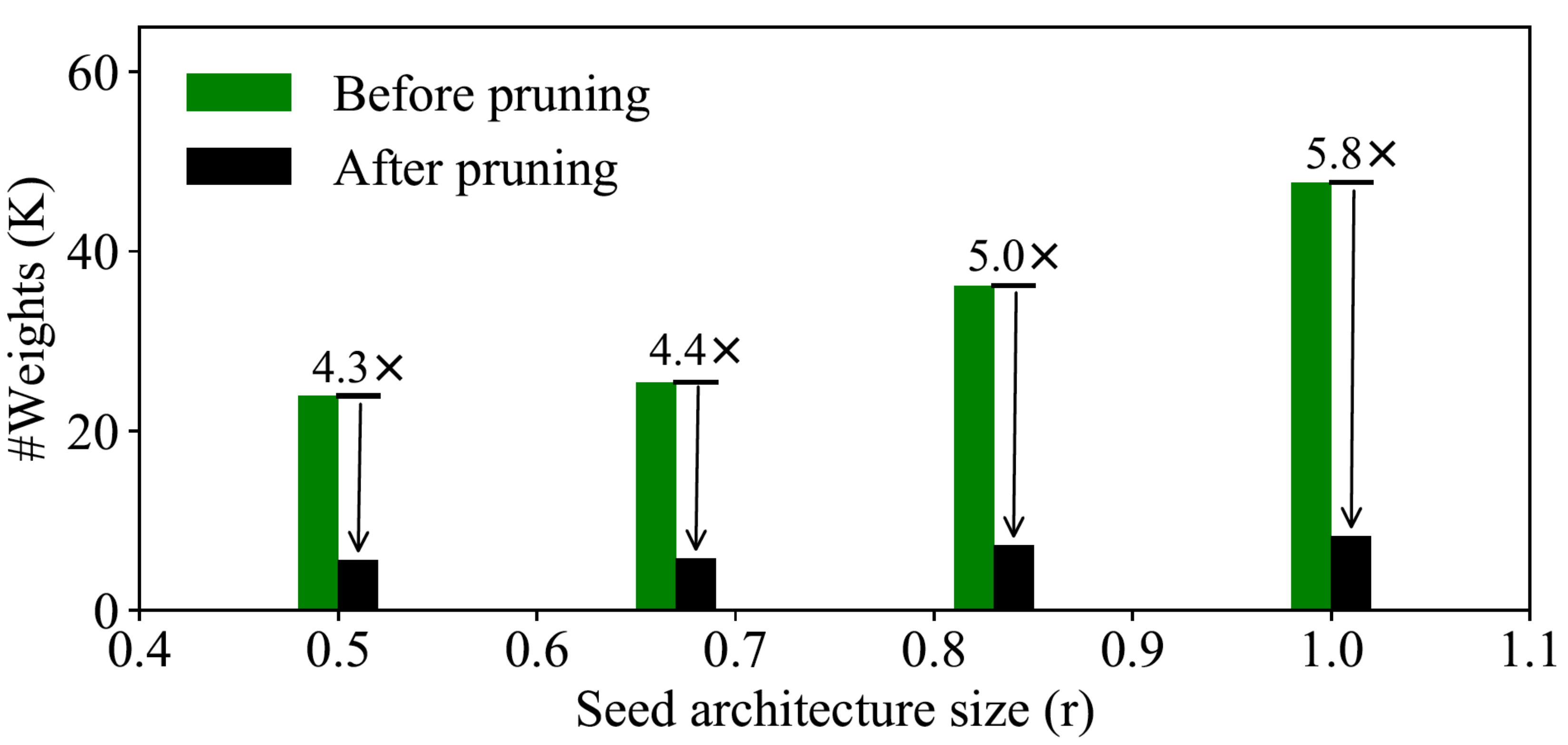}
\end {tabular}
\end {center}
\vspace{-1mm}
\caption{Compression ratio and final DNN size for different LeNet-300-100 (left) and LeNet-5 (right) seed architectures.}
\label{fig:LeNet_prune}
\vspace{-1mm}
\end{figure}

\vspace{-1mm}
\begin{enumerate}[leftmargin=6mm]\itemsep-0em 
\vspace{-1mm}
\item Smaller seed architectures often lead to smaller 
post-growth DNN sizes, but at the expense of a higher
growth time. We will later show that smaller seeds and thus smaller 
post-growth DNN sizes are better, since they also lead to smaller final
DNN sizes. 
\item When the post-growth DNN size saturates due 
to the full exploitation of the synthesis freedom for a target accuracy, a
smaller seed is no longer beneficial, as 
evident from the flat left ends of the dashed curves in Fig.~\ref{fig:LeNet_grow}. 
\end{enumerate}

\vspace{-3mm}
\subsubsection{Pruning Phase}
\vspace{-1mm}
Next, we prune the post-growth LeNet DNNs to remove their redundant
neurons/connections. We show the post-pruning DNN sizes and compression ratios
for LeNet-300-100 and LeNet-5 for the different seeds in
Fig.~\ref{fig:LeNet_prune}. 
We have two major observations for the pruning phase:
\vspace{-1mm}
\begin{enumerate}[leftmargin=6mm]\itemsep-0em 
\vspace{-1mm}
\item Larger the pre-pruning DNN, larger is its 
compression ratio. This is because larger pre-pruning DNNs have a larger
number of weights and thus also higher redundancy.
\item Larger the pre-pruning DNN, larger 
is its post-pruning DNN. Thus, to synthesize a more compact DNN, one 
should choose a smaller seed architecture (growth phase \textbf{finding 1}) 
within an appropriate range (growth phase \textbf{finding 2}).
\end{enumerate}

\vspace{-1mm}
\begin{table}[h]
\centering
\footnotesize
\caption{Different inference models for MNIST}
%\resizebox{12cm}{!}{%
\begin{tabular}{lcrrr}
\hline
Model & Method & Error & \#Param & FLOPs \\
\hline
%Linear classifier~\cite{LeNet} & - & 8.40\% & 4K & 8K \\
RBF network~\cite{LeNet} & - & 3.60\% & 794K & 1588K\\
Polynomial classifier~\cite{LeNet} & - & 3.30\% & 40K & 78K \\
K-nearest neighbors~\cite{LeNet} & - & 3.09\% & 47M & 94M \\
SVMs (reduced set)~\cite{RsSVM} & - &1.10\% & 650K & 1300K \\
Caffe model (LeNet-300-100)~\cite{caffe} & - & 1.60\% & 266K & 532K \\
LWS (LeNet-300-100)~\cite{layerwise} & Prune & 1.96\% & 4K & 8K\\
Net pruning (LeNet-300-100)~\cite{PruningHS} & Prune & 1.59\% & 22K & 43K \\
\hline
\textbf{Our LeNet-300-100: compact } & \textbf{Grow+Prune} & \textbf{1.58\%} & \textbf{3.8K} & \textbf{6.7K} \\
\textbf{Our LeNet-300-100: accurate } & \textbf{Grow+Prune} &\textbf{1.29\%} & \textbf{7.8K} & \textbf{14.9K} \\
\hline
Caffe model (LeNet-5)~\cite{caffe} & - & 0.80\% & 431K & 4586K \\
LWS (LeNet-5)~\cite{layerwise} & Prune & 1.66\% & 4K & 199K \\
Net pruning (LeNet-5)~\cite{PruningHS} & Prune &  0.77\% & 35K & 734K\\
\hline
\textbf{Our LeNet-5} & \textbf{Grow+Prune} & \textbf{0.77\%} & \textbf{5.8K} & \textbf{105K}\\
\hline
\end{tabular}
%}
\label{tab:MNIST_com}
\end{table}

\begin{table*}[h]
\centering \footnotesize
\caption{Different AlexNet and VGG-16 based inference models for ImageNet}
\resizebox{\columnwidth}{!}{%
\begin{tabular}{lcrrrr}
\hline
Model & Method & $\Delta$Top-1 err. & $\Delta$Top-5 err. & \#Param (M) & FLOPs (B) \\
\hline
Baseline AlexNet~\cite{AlexNet} & - & 0.0\% & 0.0\% & 61 ($1.0\times$) & 1.5 ($1.0\times$)\\
Data-free pruning~\cite{data_free} & Prune & +1.62\% & - &39.6 ($1.5\times$)& 1.0 ($1.5\times$)\\
Fastfood-16-AD~\cite{fastfood} & -& +0.12\% & - &16.4 ($3.7\times$)& 1.4 ($1.1\times$)\\
Memory-bounded~\cite{memory_bounded} & -& +1.62\% & - &15.2 ($4.0\times$)& - \\
SVD~\cite{svd} & -& +1.24\% & +0.83\% &11.9 ($5.1\times$) & - \\
LWS (AlexNet)~\cite{layerwise} & Prune & +0.33\% & +0.28\% &6.7 ($9.1\times$) & 0.5 ($3.0\times$)\\
Net pruning (AlexNet)~\cite{PruningHS} & Prune & -0.01\% & -0.06\% &6.7 ($9.1\times$) & 0.5 ($3.0\times$)\\
\hline
\textbf{Our AlexNet } & \textbf{Grow+Prune} &\textbf{-0.02\%} &\textbf{-0.06\%} &\textbf{3.9 ($\textbf{15.7}\times$)} & \textbf{0.33} ($\textbf{4.6}\times$) \\
\hline
Baseline VGG-16~\cite{torch_vgg} & -& 0.0\% & 0.0\% & 138  ($1.0\times$) & 30.9 ($1.0\times$)\\
LWS (VGG-16)~\cite{layerwise} & Prune& +3.61\% & +1.35\% &10.3 ($13.3\times$) & 6.5 ($4.8\times$)\\
Net pruning (VGG-16)~\cite{PruningHS} & Prune & +2.93\% & +1.26\% &10.3 ($13.3\times$) & 6.5 ($4.8\times$)\\
\hline
\textbf{Our VGG-16: accurate} & \textbf{Grow+Prune} &\textbf{-0.35\%} &\textbf{-0.31\%} &\textbf{9.9 ($\textbf{13.9}\times$)} & \textbf{6.3} ($\textbf{4.9}\times$)$^{*}$ \\
\textbf{Our VGG-16: compact} & \textbf{Grow+Prune} &\textbf{+2.31\%} &\textbf{+0.98\%} &\textbf{4.6 ($\textbf{30.2}\times$)} & \textbf{3.6} ($\textbf{8.6}\times$)$^{*}$ \\
\hline
\multicolumn{6}{l}{\scriptsize{$*$ Currently without partial-area convolution due to GPU memory limits.}}
\end{tabular}%
}
\label{tab:Alex_com}
\end{table*}

\subsubsection{Inference model comparison}  
\vspace{-1mm}
We compare our results against related results from the literature in Table~\ref{tab:MNIST_com}. Our results outperform other reference models from various design perspectives.  Without any loss of accuracy, we are able to reduce the number of connections and FLOPs of LeNet-300-100 (LeNet-5) by 70.2$\times$ ($74.3\times$) and $79.4\times$ ($43.7\times$), respectively, relative to the baseline Caffe model~\cite{caffe}.
We include the model details in the Appendix.

%\vspace{-2mm}
\subsection{AlexNet and VGG-16 on ImageNet}
\vspace{-1mm}
Next, we use NeST to synthesize DNNs for the ILSVRC 2012 image
classification dataset~\cite{imageNetdataset}.  We initialize a slim and sparse seed architecture base on the AlexNet~\cite{oneweirdtrick} and VGG-16~\cite{VGG}. 
Our seed architecture for AlexNet contains only 60, 140, 240, 210, and 160 feature maps in the five convolutional layers, and 3200, 1600, and 1000 neurons in the fully connected layers.  The seed architecture for VGG-16 uses $r=0.75$ for the first 13 convolutional layers, and has 3200, 1600, and 1000 neurons in the fully connected layers. We randomly activate 30\% of all the possible connections for both seed architectures.

Table~\ref{tab:Alex_com} compares the model synthesized by NeST with various AlexNet and VGG-16 based inference models.  We include the model details in 
the Appendix.
Our baselines are the AlexNet Caffe model (42.78\% top-1 and 19.73\% top-5 error rate)~\cite{PruningHS} and VGG-16 PyTorch model (28.41\% top-1 and 9.62\% top-5 error rate)~\cite{torch_vgg}. Our 
grow-and-prune synthesis paradigm outperforms the pruning-only 
methods listed in Table~\ref{tab:Alex_com}.  This may be explained by
the observation that pruning methods potentially inherit a certain
amount of redundancy associated with the original large DNNs.  Network growth can alleviate this phenomenon.  

Note that our current mask-based implementation of growth and pruning incurs 
a temporal memory overhead during training. If the model becomes deeper, as 
in the case of ResNet~\cite{ResNet} or DenseNet~\cite{densenet}, using masks to grow and prune 
connections/neurons/feature maps may not be economical due to this 
temporal training memory overhead. We plan to address this aspect in our 
future work.

\section{Discussions}
Our synthesis methodology incorporates three inspirations from the human brain. 

\begin{figure}[!h]
\centering
\includegraphics[width=70mm]{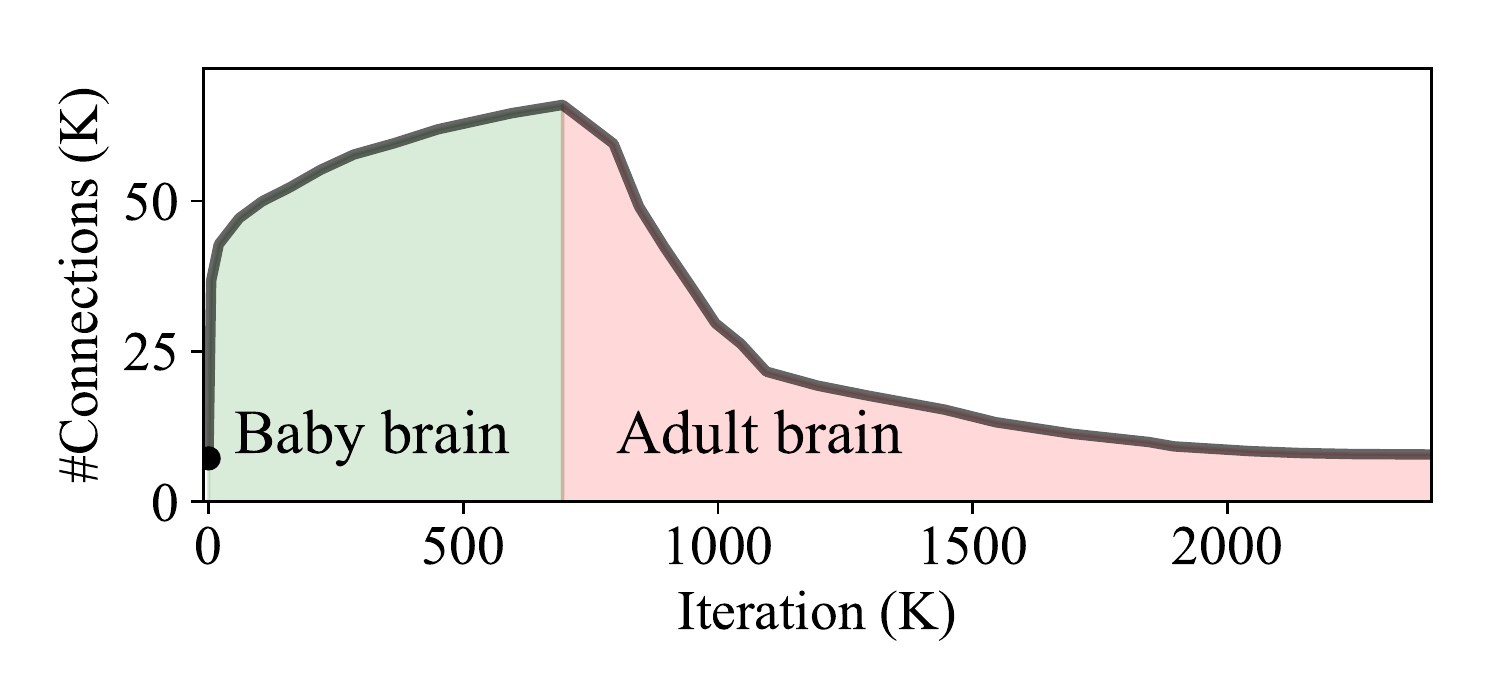}
\vspace{-2mm}
\caption{\#Connections vs. synthesis iteration for LeNet-300-100.}
\label{fig:brain}
\end{figure}

First, the number of synaptic connections in a human brain varies at different 
human ages. It rapidly increases upon 
the baby's birth, peaks after a few months, and decreases steadily thereafter. 
A DNN experiences a very similar learning process in NeST, as shown in Fig.~\ref{fig:brain}.  This 
curve shares a very similar pattern to the evolution of the number of 
synapses in the human brain~\cite{neuronnum}.

Second, most learning processes in our brain result from rewiring of 
synapses between neurons.  Our brain grows and prunes away a large amount 
(up to 40\%) of synaptic connections every day~\cite{spectrum}.  NeST wakes 
up new connections, thus effectively rewiring more neurons pairs in the 
learning process.  Thus, it mimics the `learning through rewiring' mechanism 
of human brains. 

Third, only a small fraction of neurons are active at any given time in 
human brains. This mechanism enables the human brain to operate 
at an ultra-low power (20 Watts). However, fully connected DNNs contain a
substantial amount of insignificant neuron responses per inference.  To 
address this problem, we include a magnitude-based 
pruning algorithm in NeST to remove the redundancy, thus achieving sparsity 
and compactness. This leads to huge storage and computation reductions.

\vspace{-3mm}
\section{Conclusions}
\vspace{-3mm}
In this paper, we proposed a synthesis tool, NeST, to synthesize compact 
yet accurate DNNs.  NeST starts from a sparse seed architecture, adaptively 
adjusts the architecture through gradient-based growth and magnitude-based 
pruning, and finally arrives at a compact DNN with high accuracy. For
LeNet-300-100 (LeNet-5) on MNIST, we reduced the 
number of network parameters by $70.2\times$ ($74.3\times$) and 
FLOPs by $79.4\times$ ($43.7\times$).  For AlexNet and VGG-16 on ImageNet, we reduced the
network parameters (FLOPs) by $15.7\times$ ($4.6\times$) and $30.2\times$ ($8.6\times$), respectively. 

%\section*{Acknowledgments}
%This work was supported by NSF Grant No. CNS-1617640.
%\nocite{langley00}

\bibliographystyle{IEEEtran} 
\bibliography{bibib}

\clearpage

\appendix
\Large \textbf{Appendix}
\normalsize
\section{Experimental details of LeNets}
\footnote{Our models will be released soon.}
 Table~\ref{tab:LeNet3} and 
Table~\ref{tab:LeNet5} show the smallest DNN models we could synthesize for 
LeNet-300-100 and LeNet-5, respectively. In these tables, Conv\% refers to 
the percentage of \textit{\textbf{area-of-interest}} over a full image for 
partial-area convolution, and Act\% refers to the percentage of non-zero 
activations (the average percentage of neurons with non-zero output values per inference).

\begin{table}[h]
\caption{Smallest synthesized LeNets}
\centering
\begin{minipage}[c]{0.46\textwidth}
\vspace{-0.75cm}
\subtable[LeNet-300-100 (error rate 1.29\%)]{
\begin{tabular}{lrrr}
\hline
Layer & \#Weights & Act\% & FLOPs\\
\hline
fc1 & 7032 &  46\% & 14.1K\\
fc2 & 718 &  71\% & 0.7K\\
fc3 & 94 &  100\% & 0.1K\\
\hline
Total & 7844 & N/A & 14.9K\\
\hline
\label{tab:LeNet3}
\end{tabular}
}
\end{minipage}
\begin{minipage}[c]{0.46\textwidth}
\subtable[LeNet-5 (error rate 0.77\%)]{
\begin{tabular}{lrrrr}
\hline
Layer & \#Weights & Conv\% & Act\% & FLOPs\\
\hline
conv1 & 74 & 39\% & 89\% & 45.2K\\
conv2 & 749 & 41\% & 57\% & 54.4K\\
fc1 & 4151 & N/A & 79\% & 4.7K\\
fc2 & 632 & N/A & 58\% & 1.0K\\
fc3 & 166 & N/A & 100\% & 0.2K\\
\hline
Total & 5772 & N/A & N/A & 105K\\
\hline
\label{tab:LeNet5}
\end{tabular}
}
\end{minipage}
\end{table}

\section{Experimental details of AlexNet}
Table~\ref{tab:Alex_seed} illustrates the evolution of an AlexNet seed in 
the grow-and-prune paradigm as well as the final inference model.
The AlexNet seed only contains 8.4M parameters. This number increases to
28.3M after the growth phase, and then decreases to 3.9M after the
pruning phase.  This final AlexNet-based DNN model only requires 325M FLOPs at 
a top-1 error rate of 42.76\%.

\begin{table}[h]
\centering
\caption{Synthesized AlexNet (error rate 42.76\%)}
\begin{tabular}{l|r|r|rrrrr}
\hline
Layers & \#Parameters & \#Parameters & \#Parameters & Conv\% & Act\% & FLOPs\\
\hline
& Seed & Post-Growth & \multicolumn{4}{c}{Post-Pruning}  \\
\hline
conv1 & 7K & 21K & 17K & 92\% & 87\% & 97M\\
conv2 & 65K & 209K & 107K & 91\% & 82\% & 124M\\
conv3 & 95K & 302K & 164K & 88\% & 49\% & 40M\\
conv4 & 141K & 495K & 253K & 86\% & 48\% & 36M\\
conv5 & 105K & 355K & 180K & 87\% & 56\% & 25M\\
\hline
fc1 & 5.7M & 19.9M & 1.8M & N/A & 49\% & 2.0M\\
fc2 & 1.7M & 5.3M & 0.8M & N/A & 47\% & 0.8M\\
fc3& 0.6M & 1.7M & 0.5M & N/A & 100\% & 0.5M\\
\hline
Total &8.4M & 28.3M & \textbf{3.9M} & N/A & N/A & \textbf{325M} \\
\hline
\end{tabular}
\label{tab:Alex_seed}
\end{table}

\clearpage
\section{Experimental details of VGG-16}
Table~\ref{tab:vgg_final} illustrates the details of our final compact 
inference model based on the VGG-16 architecture.  The final model only 
contains 4.6M parameters, which is 30.2$\times$ smaller than the original 
VGG-16. 

\begin{table}[h]
\caption{Synthesized VGG-16 (error rate 30.72\%)}
\centering
\begin{tabular}{lrrrrrr}
\hline
Layer & \#Param & FLOPs & \#Param & Act\% & FLOPs\\
\hline
 & \multicolumn{2}{c}{Original VGG-16}&\multicolumn{3}{c}{Synthesized VGG-16} \\
\hline
conv1\_1 & 2K & 0.2B & 1K & 64\% & 0.1B\\
conv1\_2 & 37K & 3.7B & 10K & 76\% & 0.7B\\
\hline
conv2\_1 & 74K & 1.8B & 21K & 73\% & 0.4B\\
conv2\_2 & 148K & 3.7B & 39K & 76\% & 0.7B\\
\hline
conv3\_1 & 295K & 1.8B & 79K & 53\% & 0.4B\\
conv3\_2 & 590K & 3.7B & 103K & 57\% & 0.3B\\
conv3\_3 & 590K & 3.7B & 110K & 56\% & 0.4B\\
\hline
conv4\_1 & 1M & 1.8B & 205K & 37\% & 0.2B\\
conv4\_2 & 2M & 3.7B & 335K & 37\% & 0.2B\\
conv4\_3 & 2M & 3.7B & 343K & 35\% & 0.2B\\
\hline
conv5\_1 & 2M & 925M & 350K & 33\% & 48M\\
conv5\_2 & 2M & 925M & 332K & 32\% & 43M\\
conv5\_3 & 2M & 925M & 331K & 24\% & 41M\\
\hline
fc1 & 103M & 206M & 1.6M & 38\% & 0.8M\\
fc2 & 17M & 34M & 255K & 41\% & 0.2M\\
fc3 & 4M & 8M & 444K & 100\% & 0.4M\\
\hline
Total & 138M & 30.9B & 4.6M & N/A & 3.6B\\
\hline
\label{tab:vgg_final}
\end{tabular}
\end{table}

\end{document}